# Comparative approach: Electric distribution optimization with loss minimization algorithm and particle swarm optimization

Bouabbadi Soufiane[1]

[1]Laboratory of information systems and sensors network, Hassania School of public works, Casablanca, Morocco

**Abstract:** Power systems are very large and complex, it can be influenced by many unexpected events this makes power system optimization problems difficult to solve, hence methods for solving these problems ought to be an active research topic. This review presents an overview of important mathematical comparaison of loss minimization algorithm and particle swarm optimization algorithm in terms of the performances of electric distribution.

## 1. Introduction

Electric power distribution is the final stage in the delivery of electricity. Electricity is carried from the transmission system to individual consumers. Distribution substations connects to the transmission system and lowers the transmission voltage to medium voltage ranging between 2 kV and 33 kV with the use of transformers.

Power system planning and operation offers multitudinous opportunities for optimization methods. In practice, these problems are generally large-scale, non-linear, subject to uncertainties, and combine both continuous and discrete variables.

In recent years, a number of complementary theoretical advances in addressing such problems have been obtained in the field of applied mathematics. The paper introduces a selection of these advances in the fields of non-convex optimization, in mixed- integer programming, and in optimization under uncertainty. The practical relevance of these developments for power systems planning and operation are discussed, and the opportunities for combining them, together with high-performance computing and big data infrastructures.

In this article, we are going to study the difference between the loss minimization algorithm and the particle swarm optimization in terms of performance and stability.

First, the modelisation of electric distribution will be presented. After, we are going to present the loss minimization algorithm. Later, particle swarm optimization will be advanced. Finally, we will present the results of comparaison between the performance of two methods and a brief conclusion.

## 2. Methods

*2.1. Modelisation of electric distribution*

*2.1.1. Problem description*

The problem studied is that encountered by a DNO wishing to plan the operation of its network in time, ensuring that the operational limits of its infrastructure are respected. In other terms, this amounts to finding the optimal operation, over time, of a set D of electrical devices which inject or withdraw electrical energy from the network.

*2.1.2. Network infrastructure*

The GRD infrastructure is all the electrical components of its network, namely the nodes and links (lines, cables and transformers).

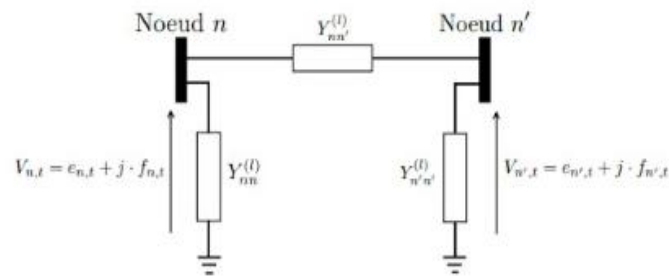

**Figure.** Representation in $\pi$ of a link connecting nodes i and j

*2.1.3. Operational limits*

The operational limits represent a set of constraints, in voltage at the nodes and in current in the links, which must be respected so as not to compromise the operation of the network.

Voltage constraint: $V_{min} \leq V_i \leq V_{max}$

Intensity limit: $I_i \leq I_{lim}$

Conservation of intensity (law of nodes).

The intensity used depends on the power.

*2.1.4. Electrical devices*

The set D of electrical devices consists of elements which are connected to nodes n ∈ N of the network and which exchange electrical energy with it. They can be differentiated into two distinct subsets:

– the set C ⊂ D of loads, they draw power from the network because they consume electrical energy.

– the set G ⊂ D of generators, they inject power into the network by producing electrical energy.

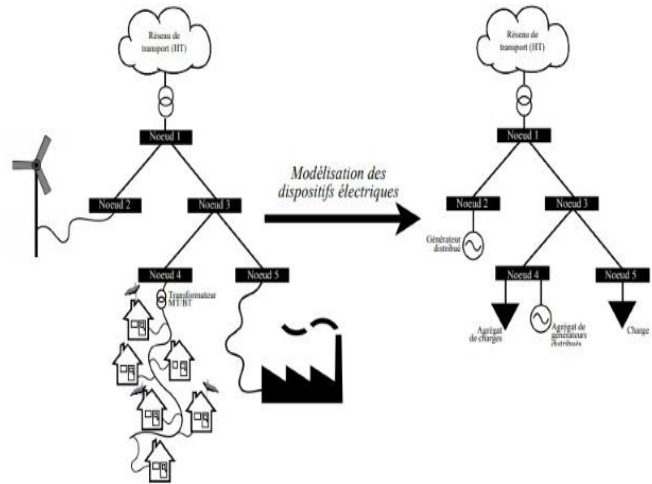

Figure. System modelisation

*2.1.5. Power losses*

Active power losses

- Power dissipated by Joule effect: $P_{Joule} = R \times I^2$

- The total power dissipated by Joule effect on an entire network is the sum of all the powers dissipated by Joule effect: $P_{Joule,tot} = P_{Joule,1} + P_{Joule,2} + \cdots + P_{Joule,5} + \cdots$

- Power of a device: $P = U \times I$

- Power dissipated by Iron effect: $P_{focault}$ , $P_{hystérésis}$

Reactive power losses

- Power dissipated by charge: $Q_{loss} = \frac{U^2}{X}$

- Reactive power compensation: $C = \frac{Q}{U^2 \times \omega}$

*2.2. Loss minimization algorithm*

*2.2.1. Structure of the Proposed Algorithm*

Although there are many loss reduction technical strategies in the current distribution network, there is little research on loss reduction optimization based on a combination of multiple types of loss reduction strategies.

The current loss reduction strategies are relatively simple and lack pertinence. Thus, a framework of combined loss reduction strategy optimization in the distribution network is proposed in this paper, which is mainly divided into three stages: weak point analysis of power loss, generation of loss reduction strategy, and combined loss reduction strategy optimization.

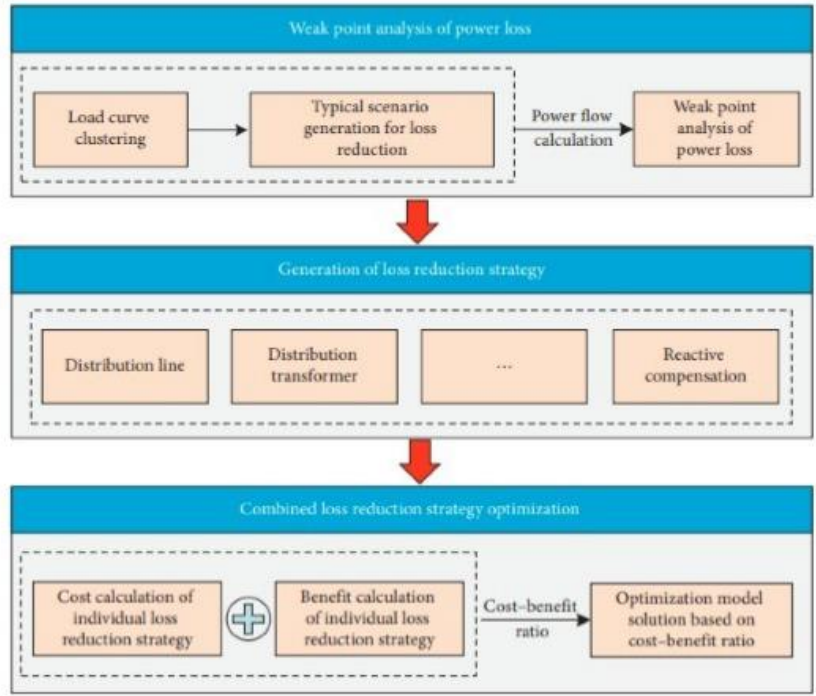

**Figure.** Loss minimization algorithm

*2.2.2. Combined Loss Reduction Strategy Optimization Model*

The loss reduction modification scheme of the distribution network is composed of different types of loss reduction strategies. Each type of loss reduction strategy has a variety of specific implementation situations for choice.

When formulating a loss reduction modification scheme, it is necessary to consider the loss reduction effect of the distribution network feeder after the loss reduction modification and to analyze the economy of the loss reduction modification.

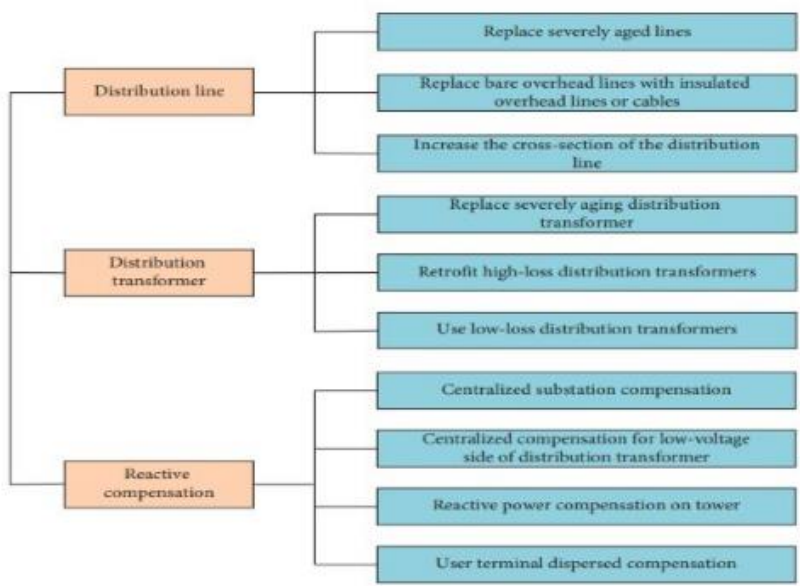

**Figure.** Strategy optimization model

*2.2.2.1. Objective Function*

This paper mainly generates loss reduction strategies from distribution transformer, distribution line, and reactive power compensation of distribution network.

Thus, the cost of power loss, the replacement cost of distribution lines, the replacement cost of distribution transformers, and the cost of reactive power compensation are needed to be considered. To optimize the comprehensive benefits of loss reduction, the objective function of the combined loss reduction strategy optimization model is established as shown in the following equation:

$$\min \quad C = C_i^{loss} + C_{vc}^{loss} + C_l^{loss} + C_t^{loss}$$

*2.2.2.2. Constraints*

- Power Loss Rate Constraint

Based on the development goals of the electric power development plan, power supply enterprises usually set the target value of the power loss rate after loss reduction modification.

$$P_{loss}\% = \frac{P_{sup} - P_{sales}}{P_{sup}} \times 100\% < \eta$$

- Power Flow Constraint

$$P_i = U_i(G_{ij}\cos\delta_{ij} + B_{ij}\sin\delta_{ij})$$

$$Q_i = U_i(G_{ij}\sin\delta_{ij} + B_{ij}\cos\delta_{ij})$$

where

$P_i$ represents the active power injected into the bus i.

$Q_i$ represents the reactive power injected to the bus i.

$U_i$ represents the voltage of bus i.

$\delta_{ij}$ denotes the phasor between bus i and j.

$G_{ij}$ denotes the conductance between bus i and j.

$B_{ij}$ represents the susceptance between bus i and j.

- Branch Transmission Capacity Constraint

The actual transmission capacity of the branch usually cannot exceed the maximum transmission capacity of the branch. In order to make the current operate within the normal range, the branch transmission capacity constraint is expressed in the following equation:

$$I_{min} \leq I_i \leq I_{max}$$

- Node Voltage Constraint

In order to make the node voltage operate within the normal range, the node voltage constraint is expressed as shown in the following equation:

$$U_{min} \leq U_i \leq U_{max}$$

- Reactive Power Compensation Capacity Constraint

The constraint of reactive power compensation capacity is shown in the following equation:

$$Q_{min} \leq Q_i \leq Q_{max}$$

*2.2.2.3. Solution Method Based on Cost-Benefit Ratio*

The cost-benefit ratio, µLR, represents the ratio of the cost of loss reduction, CLR, to the benefit of loss reduction, BLR. CLR consists of the replacement cost of distribution lines, the replacement cost of distribution transformers, and the cost of reactive power compensation. BLR is the cost corresponding to the loss reduction electricity after the loss reduction modification. It can be seen when the CLR is lower and BLR is higher, the corresponding µLR is smaller, which means that the corresponding loss reduction strategy should be selected.

$$\mu_{LR} = \frac{C_{LR}}{B_{LR}}$$

$$C_{LR} = C_{vc} + C_l + C_t$$

$$B_{LR} = C_{loss1} - C_{loss2}$$

*2.3. Particle swarm optimization*

*2.3.1. Definition*

The process of finding optimal values for the specific parameters of a given system to fulfill all design requirements while considering the lowest possible cost is referred to as an optimization. Optimization problems can be found in all fields of science.

Conventional optimization algorithms (Deterministic algorithms) have some limitations such as: Single-based solutions, Converging to local optima, Unknown search space issues.

To overcome these limitations, many scholars and researchers have developed several metaheuristics to address complex/unsolved optimization problems. Example : Particle Swarm Optimization, Grey wolf optimization, Genetic algorithm.

The Introduction to Particle Swarm Optimization (PSO) article explained the basics of stochastic optimization algorithms and explained the intuition behind particle swarm optimization (PSO).

*2.3.2. Advantages and disadvantages*

Advantages of PSO:

- Insensitive to scaling of designs variables, derivative free, very few algorithm parameters, very efficient global search algorithm and easily parallelized for concurrent processing.

Disadvantages of PSO :

- Slow convergence in the refined search stage (Weak local search ability).

*2.3.3. Flowchart model*

The proposed solution strategy can be explained with the help of flow chart. We need to maintain the voltage magnitude unchanged while minimizing the real power loss we can minimize the fuel cost via optimal adjustment of control variables.

After cost is minimized, the reactive power sub problem minimizes the total transmission loss by keeping PV bus voltage magnitude constant. Thus the total cost also decreases after second objective minimization.

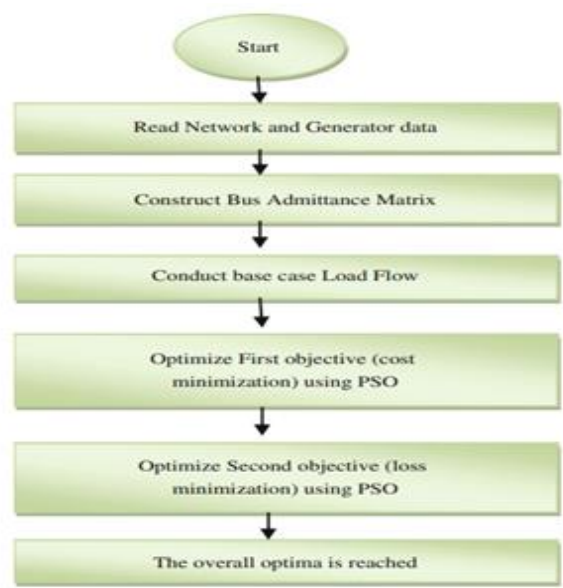

**Figure.** Flow chart of proposed method

*2.3.4. Mathematical model*

- Each particle in particle swarm optimization has an associated position, velocity, fitness value.

- Each particle keeps track of the particle_bestFitness_value particle_bestFitness_position.

- A record of global_bestFitness_position and global_bestFitness_value is maintained.

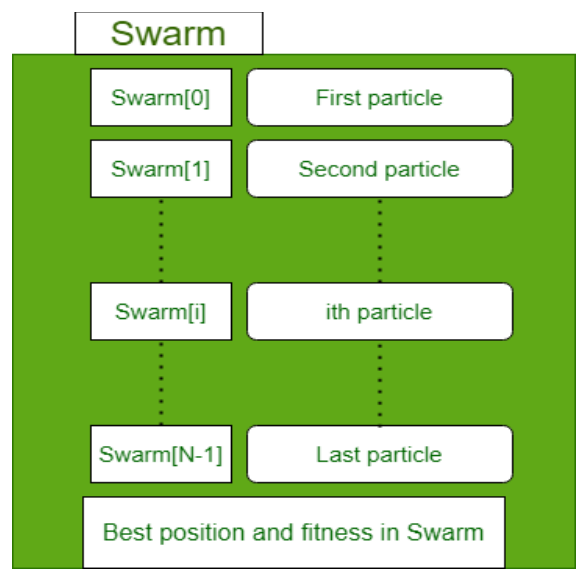

**Figure.** Data structure to store Swarm population

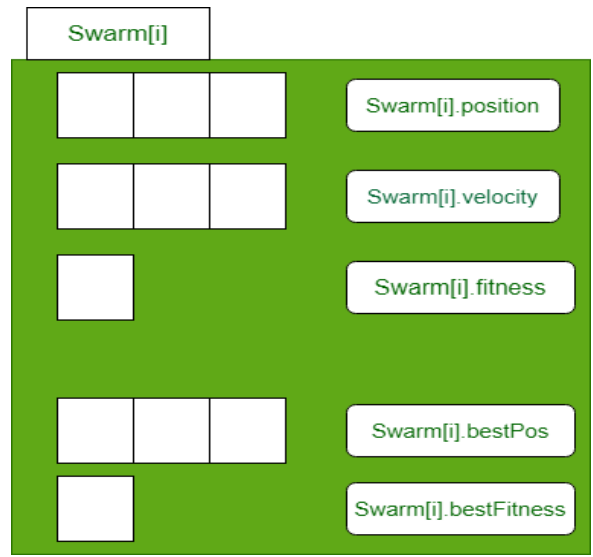

**Figure.** Data structure to store ith particle of Swarm

*2.3.5. Algorithm*

**Step1:** Randomly initialize Swarm population of N particles Xi (i=1, 2, …, n)

**Step2:** Select hyperparameter values w, c1 and c2

**Step3:** For Iter in range(max_iter):  # loop max_iter times

For i in range(N):  # for each particle

a. Compute new velocity of ith particle

swarm[i].velocity = w*swarm[i].velocity + r1*c1*(swarm[i].bestPos - swarm[i].position) + r2*c2*( best_pos_swarm - swarm[i].position)

b. Compute new position of ith particle using its new velocity

swarm[i].position += swarm[i].velocity

c. If position is not in range [minx, maxx] then clip it

if swarm[i].position < minx:

swarm[i].position = minx

elif swarm[i].position > maxx:

swarm[i].position = maxx

d. Update new best of this particle and new best of Swarm

if swaInsensitive to scaling of design variables.rm[i].fitness < swarm[i].bestFitness:

swarm[i].bestFitness = swarm[i].fitness

swarm[i].bestPos = swarm[i].position

if swarm[i].fitness < best_fitness_swarm

best_fitness_swarm = swarm[i].fitness

best_pos_swarm = swarm[i].position

End-for

End -for

**Step4:** Return best particle of Swarm

## 3. Results and discussion

### 3.1. Results

#### 3.1.1. Dataset: Profile of consumption

*Dataset summary*

To visualize the dataset, we use this code.

import pandas as pd

data= pd. read_csv ('data.txt', sep=';')

data

| | Date | Time | Global_active_power | Global_reactive_power | Voltage | Global_intensity | Sub_metering_1 | Sub_metering_2 | Sub_metering_3 |
|---|---|---|---|---|---|---|---|---|---|
| 0 | 16/12/2006 | 17:24:00 | 4.216 | 0.418 | 234.840 | 18.400 | 0.000 | 1.000 | 17.0 |
| 1 | 16/12/2006 | 17:25:00 | 5.360 | 0.436 | 233.630 | 23.000 | 0.000 | 1.000 | 16.0 |
| 2 | 16/12/2006 | 17:26:00 | 5.374 | 0.498 | 233.290 | 23.000 | 0.000 | 2.000 | 17.0 |
| 3 | 16/12/2006 | 17:27:00 | 5.388 | 0.502 | 233.740 | 23.000 | 0.000 | 1.000 | 17.0 |
| 4 | 16/12/2006 | 17:28:00 | 3.666 | 0.528 | 235.680 | 15.800 | 0.000 | 1.000 | 17.0 |
| ... | ... | ... | ... | ... | ... | ... | ... | ... | ... |
| 15918 | 27/12/2006 | 18:42:00 | 1.592 | 0.124 | 238.780 | 6.600 | 0.000 | 0.000 | 0.0 |
| 15919 | 27/12/2006 | 18:43:00 | 1.510 | 0.000 | 238.210 | 6.200 | 0.000 | 0.000 | 0.0 |
| 15920 | 27/12/2006 | 18:44:00 | 1.502 | 0.000 | 237.640 | 6.200 | 0.000 | 0.000 | 0.0 |
| 15921 | 27/12/2006 | 18:45:00 | 1.656 | 0.000 | 237.710 | 7.000 | 0.000 | 0.000 | 0.0 |
| 15922 | 27/12/2006 | 18:46:00 | 1.688 | 0.000 | 238.2 | NaN | NaN | NaN | NaN |

**Table.** Dataset

*Global active power*

To plot the global active power, we use this code.

import matplotlib. pyplot as plt

import numpy as np

e1=l1[1:20]

plt. xlabel ('Time')

plt. ylabel ('Active power')

plt. plot (e1)

plt. show ()

As a result, we obtain the following plot.

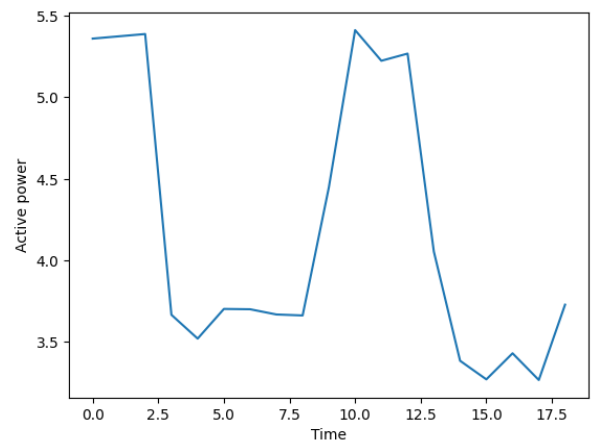

**Figure.** Active power

*Global reactive power*

To plot the global active power, we use this code.

import matplotlib. pyplot as plt

import numpy as np

e2=l2[1:20]

plt. xlabel ('Time')

plt. ylabel ('Reactive power')

plt. plot (e2)

plt. show ()

As a result, we obtain the following plot.

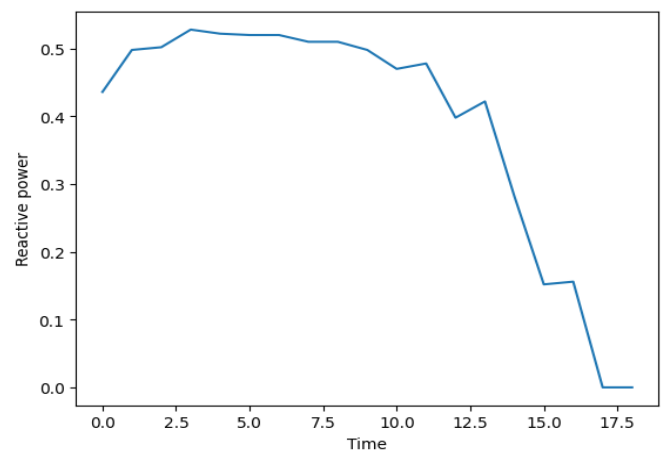

**Figure.** Reactive power

*Global intensity*

To plot the global active power, we use this code.

import matplotlib. pyplot as plt

import numpy as np

e3=l3[1:20]

plt. xlabel ('Time')

plt. ylabel ('Global intensity)

plt. plot (e3)

plt. show ()

As a result, we obtain the following plot.

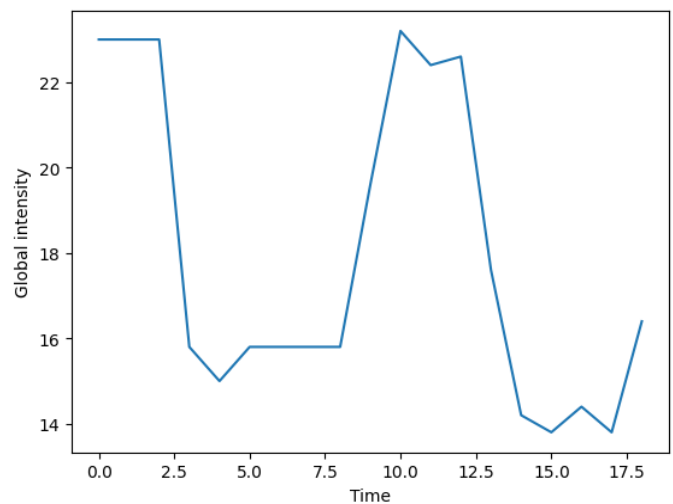

**Figure.** Global intensity

*3.1.2. Results of loss minimization algorithm*

In this algorithm, the research will turn the values of active power and reactive power to find the best values of capacitors and to show the value of yield.

When the algorithm is running based on the data of the dataset, we obtain the following results.

ULR is the cost benefit ratio to view the beneficity of electric equipments for the electric distribution.

for i in range(20):

  CLR= sum(l1[i:i+20])

  BLR= sum(l2[i:i+20])

CLR=[]

```
BLR=[]
ULR=[]
for i in range(20):
  CLR.append(sum(l1[i:i+20]))
  BLR.append(sum(l2[i:i+20]))
  ULR.append(CLR[i]/BLR[i])
```

The following is about the power lost cost.

```
import matplotlib.pyplot as plt
import numpy as np
plt.xlabel("Number")
plt.ylabel("Power Loss Cost")
plt.plot(CLR)
plt.show()
```

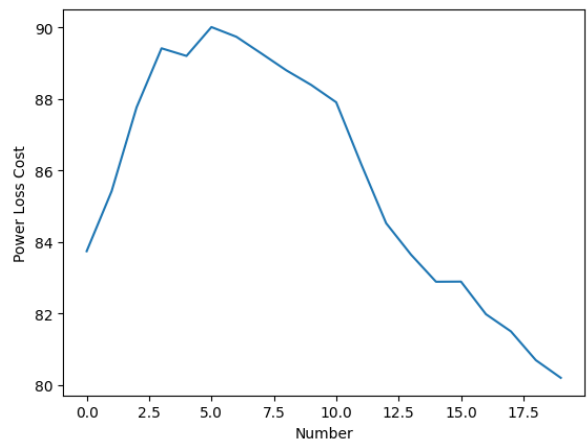

**Figure.** Power lost cost

Then, we will visualize the reactive power compensation.

```
import matplotlib.pyplot as plt
import numpy as np
plt.xlabel("Number")
plt.ylabel("Performances")
plt.plot(BLR, color='r', label='BLR')
plt.plot(CLR, color='g', label='CLR')
plt.plot(ULR, color='b', label='ULR')
```

plt.legend()

plt.show()

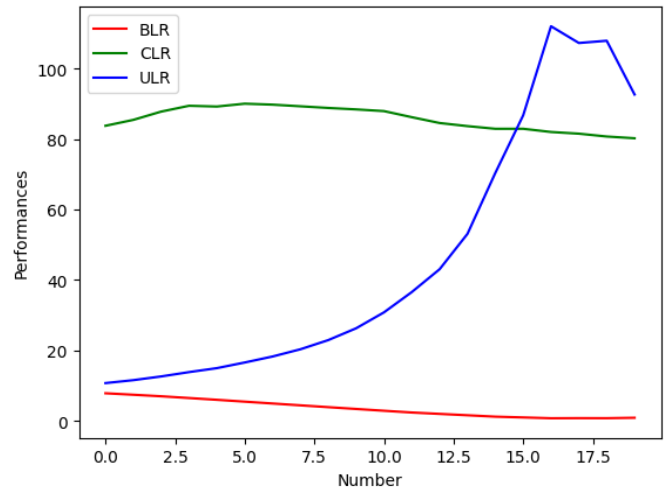

**Figure.** Performances of loss minimization algorithm

*3.1.3. Results of particle swarm optimization*

*Turning the algorithm*

In this algorithm, the research will turn the values of capacitors with swarm algorithm to find the best swarm configuration for the compesation of reactive power.

When the algorithm is running based on the data of the dataset, we obtain the following results.

Fitness is the ability of the data value of capacitors for active power.

Begin particle swarm optimization on rastrigin function

Goal is to minimize Rastrigin's function in 200 variables Function has known min = 0.0 at (Setting num_particles = 10 Setting max_iter = 100

Starting PSO algorithm

Iter = 5 best fitness = 195.882

Iter = 10 best fitness = 176.540

Iter = 15 best fitness = 169.292

Iter = 20 best fitness = 169.292

Iter = 25 best fitness = 169.292

Iter = 30 best fitness = 165.916

Iter = 35 best fitness = 165.916

Iter = 40 best fitness = 162.998

Iter = 45 best fitness = 160.412

Iter = 50 best fitness = 158.921

Iter = 55 best fitness = 156.970

Iter = 60 best fitness = 146.125

Iter = 65 best fitness = 145.721

Iter = 70 best fitness = 140.158

Iter = 75 best fitness = 126.345

Iter = 80 best fitness = 123.098

Iter = 85 best fitness = 119.881

Iter = 90 best fitness = 111.329

Iter = 95 best fitness = 110.468

PSO completed

Best solution found: ['0.048813', '0.950382', '-0.080204', '0.072667', '0.197269', '0.003676', '-0.100668', '-0.027871', '0.919538', '0.065172', '0.094082', '-0.127899', '0.040985', '0.050443', '-0.097652', '0.040515', '0.044928', '0.000075', '0.000000', '0.000000', '0.000000', '0.000000', '0.000000', '0.000000', '0.000000', '0.000000', '0.000000', '0.000000', '0.000000', '0.000000', '0.000000', '0.000000', '0.000000', '0.000742', '-0.012201', '0.000000', '0.063907', '0.046526', '0.070804', '-0.075356', '0.008491', '0.028560', '-0.020775', '-0.055102', '0.138241', '0.083133', '0.012301', '-0.170155', '0.074332', '-0.004839', '0.105315', '0.083135', '-0.117491', '-0.016693', '0.004117', '0.071149', '0.000000', '0.000000', '0.000000', '0.000000', '0.000000', '0.000000', '0.000000', '0.000000', '0.000000', '0.000000', '0.000000', '0.000000', '0.017794', '0.042155', '-0.040874', '0.012950', '0.000000', '0.000000', '0.000000', '0.003064', '0.052019', '-0.011069', '-0.011891', '-0.001604', '-0.012865', '0.122345', '0.021783', '-0.012144', '0.075668', '-0.018821', '-0.012039', '0.079587', '-0.053287', '0.000000', '0.000000', '0.000000', '0.082137', '-0.012242', '0.098231', '0.090317', '0.020860', '0.037082', '0.055033', '0.032382', '0.024991', '0.000000', '0.072250', '-0.074181', '0.053561', '0.093728', '0.012936', '0.000000', '0.000000', '0.000000', '0.000000', '0.000000', '0.000000', '0.000000', '0.000000', '0.000000', '0.000000', '0.019869', '0.000000', '0.003116', '0.000000', '0.020165', '-0.046963', '0.034438', '0.005342', '0.114528', '0.013943', '0.090651', '0.067123', '0.052181', '-0.017627', '0.009294', '0.000000', '0.000000', '0.057472', '-0.064115', '-0.021365', '0.213388', '0.022634', '0.211897', '-0.022396', '0.054950', '-0.055599', '-0.029585', '0.010559', '-0.067792', '-0.015447', '0.055512', '-0.096250', '0.058651', '0.000000', '0.041976', '-0.001659', '-0.010375', '0.045574', '0.007113', '-0.015774', '0.009503', '0.038393', '-0.033626', '0.028512', '0.000865', '0.000000', '0.000000', '0.000000', '0.000000', '0.006321', '-0.057083', '0.042169', '0.031496', '0.146471', '0.032164', '0.024358', '0.000000', '0.000000', '0.000000', '0.000000', '0.000000', '0.000000', '0.000000', '0.000000', '0.000000', '-0.005161', '0.053367', '0.038673', '0.006413', '0.037749', '0.000913', '0.016967', '-

0.032712', '0.005920', '0.026302', '0.036561', '0.096994', '0.007373', '0.022419', '0.113608', '-0.025637', '0.005475', '0.025987', '-0.021156']

Fitness of best solution = 107.060385

End particle swarm for rastrigin function

The code is completd witouth any errors with the best swarm configuration.

Fitness is the ability of the data value of capacitors for the compensation of reactive power.

Begin particle swarm optimization on rastrigin function

Goal is to minimize Rastrigin's function in 200 variables Function has known min = 0.0 at (Setting num_particles = 10 Setting max_iter = 100

Starting PSO algorithm

Iter = 5 best fitness = 2599.217

Iter = 10 best fitness = 2599.217

Iter = 15 best fitness = 2599.217

Iter = 20 best fitness = 2599.217

Iter = 25 best fitness = 2570.804

Iter = 30 best fitness = 2570.804

Iter = 35 best fitness = 2570.804

Iter = 40 best fitness = 2511.640

Iter = 45 best fitness = 2447.251

Iter = 50 best fitness = 2447.251

Iter = 55 best fitness = 2354.132

Iter = 60 best fitness = 2282.404

Iter = 65 best fitness = 2263.765

Iter = 70 best fitness = 2261.912

Iter = 75 best fitness = 2261.912

Iter = 80 best fitness = 2198.497

Iter = 85 best fitness = 2198.497

Iter = 90 best fitness = 2198.063

Iter = 95 best fitness = 2194.678

PSO completed

Best solution found: ['-2.183467', '0.953548', '-1.391651', '3.265416', '2.391669', '-1.864360', '-0.604359', '-2.877062', '-2.301748', '3.842722', '-0.222927', '0.669589', '-0.781776', '1.076682', '-2.221679', '-1.130106', '-2.003562', '-2.458907', '-0.506874', '-0.816786', '0.012080', '-1.110548', '1.896282', '1.863231', '0.829427', '0.093193', '-0.128719', '0.893509', '-0.895500', '0.002228', '0.178787', '-0.162760', '-2.159304', '0.231979', '-2.620145', '1.695631', '0.663011', '-1.048289', '-2.440579', '2.033472', '4.203047', '1.918726', '6.824284', '-0.689263', '1.167084', '-5.000215', '4.670019', '1.990537', '2.073411', '1.660184', '-0.909954', '-2.645651', '-0.332943', '-2.839466', '2.219205', '1.923027', '-2.642269', '1.083379', '3.816709', '-0.187519', '1.025116', '1.250896', '4.331597', '2.906166', '-0.226219', '0.690826', '0.469869', '1.742865', '0.131248', '-2.039961', '-4.012813', '0.296230', '0.950066', '-1.060328', '-0.642453', '1.840690', '0.185719', '1.013282', '0.135256', '0.730176', '1.300194', '-0.088173', '-3.761053', '1.371650', '0.114503', '1.001887', '1.966248', '-2.053022', '0.346632', '-0.655002', '-0.109072', '1.788866', '-1.500318', '-1.946222', '1.186794', '0.608686', '1.159922', '1.097523', '2.308782', '-0.356571', '1.731363', '-2.454526', '0.855709', '1.958047', '0.137743', '3.672530', '-0.143978', '1.998084', '-0.933176', '1.195352', '4.671388', '0.642276', '3.273142', '0.840626', '0.002979', '-0.958490', '1.775999', '-0.947570', '0.144036', '0.890622', '1.747463', '0.199109', '2.764243', '-4.273229', '3.769598', '0.751971', '0.999755', '1.748799', '0.237950', '-1.023727', '1.733650', '0.906336', '0.908646', '1.035962', '-0.987321', '0.950849', '3.777176', '-1.459373', '0.167275', '0.151835', '2.746764', '-3.202045', '0.429886', '-4.304787', '-1.169802', '-0.183836', '1.448848', '-1.998550', '0.639413', '-0.713794', '1.122164', '1.330402', '0.187407', '1.929653', '0.823875', '0.923578', '-0.921660', '0.045625', '0.279938', '0.502168', '0.076659', '1.831992', '-1.788151', '2.289143', '0.989936', '1.051748', '2.358462', '0.588742', '-1.788180', '0.158368', '1.617777', '1.167882', '0.919429', '-0.063166', '1.366075', '-1.466239', '1.353893', '-0.650562', '-0.183135', '3.950612', '3.982958', '0.011032', '2.913098', '-0.388262', '-1.078812', '0.337668', '-3.907618', '2.077140', '-1.812489', '-3.895097', '1.097816', '1.458317', '1.871343', '2.794766', '-0.046465', '2.022763', '-1.040607', '-1.941625', '1.939860', '1.052197']

Fitness of best solution = 2160.720835

End particle swarm for rastrigin function

The code is completd witouth any errors with the best swarm configuration.

*Visualizing the results*

The fitness of the swarm of our datest is declining when augmenting the iterations.

We found the following plot.

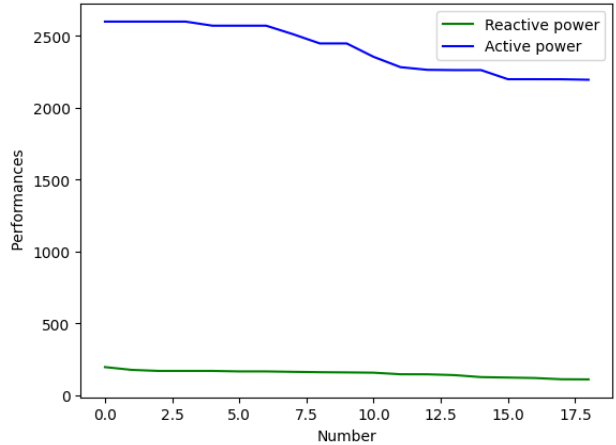

**Figure**. Fitness of the swarm

The best solution is visualizing in the following graph.

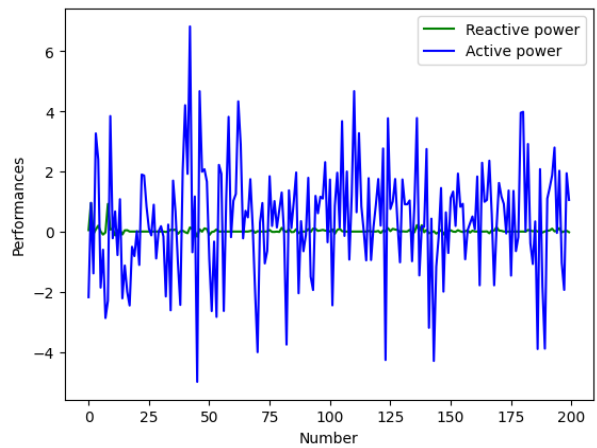

**Figure**. Swarm best solution

The figure is visualizing the capacitors value value in order to find the best value for the compensation of reactive power.

*3.2. Discussion*

Concerning loss minimization algorithm, the power loss algorithm is declining when applying the iterations due to the consumption of power.

The cost benefit ratio is very high because the reactive power compensation is in the norm.

Concerning the particle swarm optimization, the active power is more fluctuating than the reactive power when applying the algorithm.

In addition, the active power is higher than the reactive power.

When comparing the two algorithm, the loss minimization algorithm is more performing than the particle swarm algorithm, as we can see in the figure.

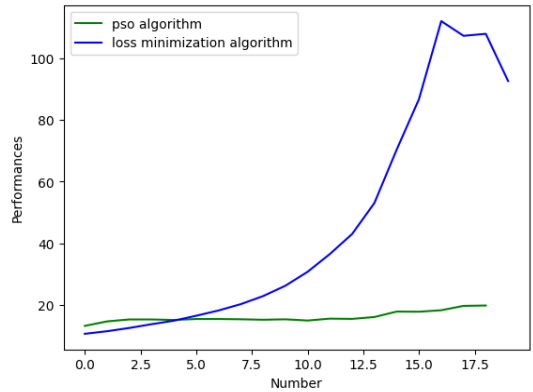

**Figure.** Comparaison of cost benefit ratio

## 4. Conclusion

The electric distribution optimization is a serious problem that can be treated with different techniques such as loss minimization algorithm and particle swarm optimization.

In this article, we make a comparaison between the two algorithms.

Power loss minimization is an algorithm that compensates the reactive power and minimizes the lost in active power.

Particle swarm optimization is a method that tried to optimize the electric distribution considering the final consumers as a swarm and tried to optimize the weights of the swarm.

Comparing the two algorithms, we find that the loss minimization is more performing than the particle swarm optimization regarding the performances of electric distribution (cost benefit ratio).